\documentclass[10pt,twocolumn,letterpaper]{article}
\usepackage{times}
\usepackage{epsfig}
\usepackage{graphicx}
\usepackage{amsmath}
\usepackage{amssymb}
\usepackage{booktabs}
\usepackage{soul}
\usepackage{enumitem}
\usepackage{bbm}
\usepackage{mathtools}
\usepackage{xurl}
\usepackage{diagbox}
\usepackage{multirow}
\usepackage[normalem]{ulem}
\usepackage[table,xcdraw]{xcolor}
\usepackage{stfloats}
\usepackage{pifont}% http://ctan.org/pkg/pifont

\newcommand{\cmark}{\ding{51}}%
\newcommand{\xmark}{\ding{55}}%
\useunder{\uline}{\ul}{}
\DeclareMathOperator*{\argmax}{arg\,max}

\usepackage[style=ieee]{biblatex}
\usepackage[breaklinks=true,bookmarks=false]{hyperref}

\begin{document}

%%%%%%%%% TITLE
\title{Real-Time Driver Monitoring Systems through Modality and View Analysis}

\author{Yiming Ma\\
University of Warwick\\
Coventry, UK\\
% {\tt\small yiming.ma.1@warwick.ac.uk}
\and
Victor Sanchez\\
University of Warwick\\
Coventry, UK\\
% {\tt\small v.f.sanchez-silva@warwick.ac.uk}
\and
Soodeh Nikan\\
Ford Motor Company\\
USA\\
% {\tt\small snikan@ford.com}
\and
Devesh Upadhyay\\
Ford Motor Company\\
USA\\
% {\tt\small dupadhya@ford.com}
\and
Bhushan Atote\\
University of Warwick\\
Coventry, UK\\
% {\tt\small bhushan.atote@warwick.ac.uk}
\and
Tanaya Guha\\
University of Glasgow\\
Glasgow, UK\\
% {\tt\small tanaya.guha@glasgow.ac.uk}
}
\maketitle

\thispagestyle{empty}

%%%%%%%%% ABSTRACT
\begin{abstract}
Driver distractions are known to be the dominant cause of road accidents. While monitoring systems can detect non-driving-related activities and facilitate reducing the risks, they must be accurate and efficient to be applicable. Unfortunately, state-of-the-art methods prioritize accuracy while ignoring latency because they leverage cross-view and multimodal videos in which consecutive frames are highly similar. Thus, in this paper, we pursue time-effective detection models by neglecting the temporal relation between video frames and investigate the importance of each sensing modality in detecting drives' activities. Experiments demonstrate that 1) our proposed algorithms are real-time and can achieve similar performances (97.5\% AUC-PR) with significantly reduced computation compared with video-based models; 2) the top view with the infrared channel is more informative than any other single modality. Furthermore, we enhance the DAD dataset by manually annotating its test set to enable multiclassification. We also thoroughly analyze the influence of visual sensor types and their placements on the prediction of each class. The code and the new labels will be released.
% Driver state monitoring (DSM) is a critical component in reducing driving risks, enhancing passenger safety and achieveing autonomy for cars. DSM is a task of identifying various driving and non-driving activities, which often use visual sensors of different modalities (e.g., depth, Infrared) providing multiple views (front, side). This multimodal, mutiview data adds high computing load to a DSM system, which needs to operate in real-time (while being accurate) to be fully effective. Existing DSM methods only aim to maximize accuracy ignoring two practical aspects:real-time inference, and which sensors and which views are the most important. This paper presents an efficient DSM that can be implemneted real-time while investigating the contribution of different sensing modalities and their views. Our proposed DSM model is based on images for efficiency, and yet, can achieve the same level of performance as the video-based DSM models with significantly reduced computation. We show that the top view with the infrared sensor is the most informative channel for DSM than any other single modality on the DAD dataset. Furthermore, we enhance the DAD dataset by manually annotating its test set (previously had binary lables) for various driver states. Using these new labels, we bring new insights on how different visual sensors and their views influence the detection of different drivers states. The new labels and the code will be released shortly.
\end{abstract}

%%%%%%%%% BODY TEXT
\section{Introduction}
\label{sec:1}

While the proliferation of on-road traffic has dramatically benefited society, it has also increased fatal traffic accidents. According to the World Health Organization, until June 2022, there have been about 1.3 million casualties and more than 20 million injuries from car crashes every year, and these accidents cost most countries approximately 3\% of their GDPs. In these crashes, one of the dominant contributing factors is human errors from inattention, such as preoccupation with mobile phones while driving. Hence, for L2+ self-driving-enabled cars, it is crucial to develop effective \textit{driver monitoring systems} (DMSs) to estimate the drivers' readiness for driving and take over the control when necessary to prevent accidents.

As an essential information source, vision is often exploited by DMSs to detect drivers' \textit{non-driving-related activities} (NDRAs). DAD \cite{kopuklu2021driver} is one of the latest video databases for vision-based monitoring systems. In addition to the rich diversity of NDRAs in the training set, its test set also contains unseen types of actions. Since there can be unboundedly many actions that drivers may conduct and the unseen cases in the test set of DAD enable it to better generalize to realistic driving, we decide to establish our work on this open-set recognition dataset. However, the test-set labels of DAD are binary, only indicating whether drivers are participating in NDRAs or not. This inadequacy hinders the multiclassification of drivers' activities, which is essential since different activities may not be equally hazardous. Hence, we manually annotate the test set with the specific task names to allow the recognition of these actions and class-based evaluation of DMSs.

On the other hand, the latest vision-based DMSs are not sufficiently efficient. In-cockpit vision systems (including DAD) usually consist of visual sensors of various types (e.g. RGB) installed at different locations (e.g. top) to provide videos of the driver from diverse streams and views. Thus, to maximize the accuracy of activity detection, existing approaches often employ spatial and temporal information from all modalities and views. These methods unavoidably introduce billions of \textit{floating-point operations} (FLOPs) into inference, which embedded devices cannot perform in real time. Therefore, in this paper, we aim to develop a real-time DMS by neglecting the temporal dimension and only employing the most informative vision source. Experiments on DAD demonstrate that 1) neighboring frames within a sequence are highly resembling; 2) single-modal architectures based on the top view and the infrared modality can also achieve state-of-the-art performances. These two findings validate our approach. 

\begin{table*}[hbtp]
\begin{tabular}{l | l l l}
\hline
\textbf{NDRAs in DAD training set} & \multicolumn{3}{c}{\textbf{NDRA labels we annotated for DAD test set}} \\
\hline
Talking on the phone - left         & Talking on the phone - left  & \textcolor{red}{Adjusting side mirror}      & \textcolor{red}{Wearing glasses}      \\
Talking on the phone - right        & Talking on the phone - right & \textcolor{red}{Adjusting clothes}          & \textcolor{red}{Taking off glasses}   \\
Messaging left                    & Messaging left             & \textcolor{red}{Adjusting glasses}          & \textcolor{red}{Picking up something} \\
Messaging right                   & Messaging right            & \textcolor{red}{Adjusting rear-view mirror} & \textcolor{red}{Wiping sweat}         \\
Talking with passengers           & Talking with passengers    & \textcolor{red}{Adjusting sunroof}          & \textcolor{red}{Touching face/hair}   \\
Reaching behind                   & Reaching behind            & \textcolor{red}{Wiping nose}                & \textcolor{red}{Sneezing}             \\
Adjusting radio                   & Adjusting radio            & \textcolor{red}{Head dropping (dozing off)} & \textcolor{red}{Coughing}             \\
Drinking                          & Drinking                   & \textcolor{red}{Eating}                     & \textcolor{red}{Reading} \\
\hline
\end{tabular}
\caption{Details of the \textit{non-driving-related activities} ({NDRA}s) in the DAD dataset \cite{kopuklu2021driver}. Activities in \textcolor{black}{black} are present in both the training and test sets, and those in \textcolor{red}{red} are exclusive to the test set. The original DAD dataset only provides binary (normal/anomalous) labels for the test set as it is initially developed for anomaly detection.}
\label{tbl:train_test_ndras}
\end{table*}

Our contributions are summarized as follows:
\begin{enumerate}[itemsep=2pt,topsep=0pt,parsep=0pt]
    \item We propose efficient image-based DMSs with comparable performances (97.5\% AUC-PR \& 95.6\% AUC-ROC) with state-of-the-art video-based models. Unlike other methods' substantial computation load, our models' low latency makes them deployable in the real world.
    \item We analyze the performance of our models on each view and modality and provide the most economical solution to the placement (top) and the type (IR) of cameras that DMSs should leverage.
    \item We annotate the test set of DAD with specific activities, thereby enabling the multiclassification of drivers' actions on it. This particularized labelling allows detailed analysis of algorithms based on each class, and detecting the most dangerous distractions (e.g. using mobile phones) can be prioritized.
\end{enumerate}

\section{Related Work}
\label{sec:2}

\subsection{In-Car Vision-Based Datasets}
\label{sec:2.1}

Early datasets focus only on parts of the driver's body, like the head \cite{diaz2016reduced, massoz2016ulg, roth2019dd, schwarz2017driveahead, weng2016driver} or hands \cite{das2015performance, kopuklu2020drivermhg, ohn2013driver}. Although these may have contributory values in other tasks, such as gesture recognition and pose study, only a narrow range of the whole-body movements are typically captured, thereby covering very few NDRAs. 

Later datasets that followed \cite{abouelnaga2018real, martin2019drive,  ortega2020dmd} instead concentrate on body actions. AUC-DD \cite{abouelnaga2018real} is based on a single sensing modality, with images collected from a side view perspective and the RGB stream. By contrast, DMD \cite{ortega2020dmd} is a multimodal and video-based dataset, currently the largest for SAE L2-L3 autonomous driving. It consists of 41 hours of videos recorded from three different views and three distinct channels. However, its training set and test set share the same classes of activities, so models trained on it may be overfitted to these seen actions and thus fail to generalize well in real-world scenarios. In comparison, DAD \cite{kopuklu2021driver} is devised for open-set recognition. In addition to the categories in the training set, its test set also comprises unseen types of activities, as illustrated in Table \ref{tbl:train_test_ndras}. Therefore, this database is more appropriate for estimating DMSs' real-world performances. However, the DAD test labels only indicate whether or not a driver is engaging in driving or NDRAs, without specifying the categories of actions. This binary labelling confines its usage to anomaly detection. As a result, DMSs developed using this dataset remain oblivious to the type of potentially distracting activity. This granularity may be critical in designing a composite attention metric where the different types of NDRAs may have different sensitivities. Hence, we found that a richer set of labels is essential for studying NDRAs and their impacts on overall driver attention.

\begin{figure}[htp]
\begin{minipage}[t]{0.49\linewidth}
    \centering
    \centerline{\includegraphics[width=4.0cm]{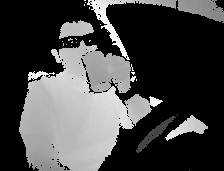}}
    \centerline{(a) Front \& Depth.}\medskip
\end{minipage}
\hfill
\begin{minipage}[t]{0.49\linewidth}
    \centering
    \centerline{\includegraphics[width=4.0cm]{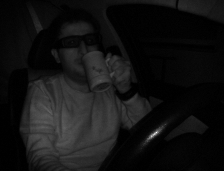}}
    \centerline{(b) Front \& IR.} \medskip
\end{minipage}
\begin{minipage}[t]{0.49\linewidth}
    \centering
    \centerline{\includegraphics[width=4.0cm]{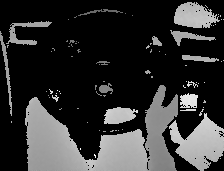}}
    \centerline{(c) Top \& Depth.}\medskip
\end{minipage}
\hfill
\begin{minipage}[t]{0.49\linewidth}
    \centering
    \centerline{\includegraphics[width=4.0cm]{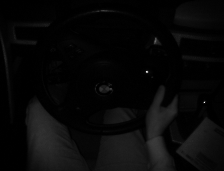}}
    \centerline{(d) Top \& IR.} \medskip
\end{minipage}
\caption{During the collection of DAD \cite{kopuklu2021driver}, two cameras with synchronized depth and IR modalities were installed at the top and front of the cabin, capturing the movement of hands and parts of the body, respectively. The resolution of each video frame is $224 \times 171$.}
\label{fig:dad_samples}
\end{figure}

\subsection{Vision-Based Driver Monitoring Systems}
\label{sec:2.2}

\noindent\textbf{Image-based DMSs.} \cite{abouelnaga2018real} proposes an approach based on object detection, in which the hand and face areas of the driver in the image are detected first and then fed into an ensemble of \textit{convolutional neural networks} (CNNs). Established on this architecture, a model proposed in \cite{eraqi2019driver} further includes a skin segmentation branch to resolve the issue of variable lighting conditions that imposes performance degradation onto RGB-based models. Contrary to multi-branch structures, \cite{baheti2018detection} leverages classic CNN classifiers such as VGG \cite{simonyan2015very}, also leading to state-of-the-art results. This finding motivates us to establish our work on pre-trained ResNets \cite{he2016deep} and MobileNet \cite{sandler2018mobilenetv2}.

\noindent\textbf{Video-based DMSs.} Currently, most video-based methods \cite{kopuklu2021driver, ortega2020dmd} leverage 3D CNN classifiers pre-trained on Kinetics-600 \cite{carreira2017quo, kopuklu2019resource}. \cite{ortega2020dmd} proposes a DMS on the DMD dataset \cite{ortega2020dmd} that first exploits 3D CNNs (MobileNet-V2 \cite{sandler2018mobilenetv2} and ShuffleNet-V2 \cite{ma2018shufflenet}) as the backbone to extract spatial-temporal features and then uses a consensus module to further capture the temporal correlations between frames. 3D ResNet-18 \cite{he2016deep}, 3D MobileNet-V2 \cite{sandler2018mobilenetv2} and 3D ShuffleNet-V2 \cite{ma2018shufflenet} are utilised in \cite{kopuklu2021driver} to extract embeddings and contrastive learning with noise estimation \cite{gutmann2010noise} is adopted to optimize the cosine similarities of these representations. However, \cite{tran2018closer} suggests that spatial details are more influential than temporal relations. We also prove that high-level similarities exist between consecutive frames. These two observations indicate that leveraging all frames within a video clip may not be cost-effective, so an image-based DMS does not need to make inferences for each frame, thereby showing the potential of real-time performances.

\subsection{Multimodal Feature Fusion}
\label{sec:2.3}

The fundamental problem in multimodality is fusion. Previous DMSs with multimodality \cite{kopuklu2021driver, ortega2020dmd} are based on decision-level fusion, while combing multimodal features is rarely studied. We regard multimodal feature fusion as a particular case of general feature fusion, and various approaches have been proposed during its evolvement. Earlier ones leverage linear operations such as concatenation (e.g. Inception \cite{szegedy2015going, szegedy2016rethinking, szegedy2017inception}) and addition (e.g. ResNet \cite{he2016deep}, FPN \cite{lin2017feature}, U-Net \cite{ronneberger2015u}). This rigid linear aggregation treats all features equally, thereby lacking adaptation. Thus, later works, such as SENet \cite{hu2018squeeze} and SKNet \cite{li2019selective}, employ attention mechanisms to enable weighted fusion. However, as shown in \cite{dai2021attentional}, these two methods have several drawbacks (like a naive initial integration) that can hurt models' performances. Hence, to address these issues \cite{dai2021attentional} proposes AFF and iAFF, which assess the importance of each feature from both a global and a local perspective, and thus they can adapt to various scenarios. Nevertheless, the original \textit{multi-scale feature fusion module} (MS-CAM) within AFF and iAFF only supports the simultaneous fusion of two features. To resolve this issue, we extend the number of attention heads to accommodate more features.

\section{Methodology}
\label{sec:3}

\subsection{Data Labelling}
\label{sec:3.1}

\begin{table}[tp]
\resizebox{\linewidth}{!}{
\begin{tabular}{llllll}
\hline
\textbf{Video ID} & \textbf{Frames} & \textbf{Original Labels} & \textbf{New Labels} \\ \hline
rec1     & 0 - 164       & Anomalous       & Adjusting Radio\\
rec1     & 165 - 513       & Normal          & Normal Driving\\
rec1     & 514 - 1150      & Anomalous       & Drinking\\
rec1     & 1151 - 1831      & Anomalous       & Normal Driving\\
rec1     & 1832 - 2336      & Anomalous       & Adjusting side mirror\\
rec1     & 2337 - 2886      & Anomalous       & Normal Driving\\
rec1     & 2887 - 3688      & Anomalous       & Reading\\
rec1     & 3689 - 4399      & Anomalous       & Normal Driving                     \\ 
\hline
\end{tabular}}
\caption{Examples of the new labels of the test set of DAD \cite{kopuklu2021driver}. The original version does not particularize the types of NDRAs, so we review each video clip to label it. This more detailed version allows multiclassification on DAD and class-based evaluation.}
\label{tbl:orig_labels}
\end{table}

\noindent\textbf{Annotating the test set.} The original test-set labels of DAD \cite{kopuklu2021driver} take only two values -- ``anomalous'' and ``normal'', denoting whether or not the driver is engaging in NDRAs. We manually scrutinize each video clip (from both views and both modalities) and label it with the corresponding class (see Table \ref{tbl:train_test_ndras}). During this process, we find that some activities, such as adjusting the radio, can only be confirmed from the top view, while other tasks, like talking to passengers, are only observable from the front view. This discovery demonstrates that even for human annotators, both views are required to determine the driver's state, so cross-view models have the potential to perform better. We also find that in addition to those NDRAs listed in Table \ref{tbl:train_test_ndras}, there are two new tasks -- ``looking for something'' and ``yawning'', so in fact,  there exist 26 different categories of NDRAs in the test set. Table \ref{tbl:orig_labels} shows a slice of our annotations.

\noindent\textbf{Labels in training and testing.} In the test set, there are video clips in which drivers suddenly switch to the other hand to hold their phones. Therefore, we neglect the effect of different hand involvement to allow training and testing models effortlessly. To be more specific, we merge ``messaging left'' and ``messaging right'' into ``messaging'' and ``talking on the phone - left'' and ``talking on the phone - right'' into ``talking on the phone''. As for unseen NDRAs, we combine all of them into a new class, ``unseen''. The convention is illustrated in Table \ref{tbl:new_labels}.

\begin{table}[tp]
\resizebox{\linewidth}{!}{
\begin{tabular}{| l | l | c | c |}
\hline
\textbf{Activities in Training Set} & \textbf{Activities in Test Set}  & \textbf{NDRA?} & \textbf{Notation} \\
\hline
Normal driving  & Normal driving & \xmark & $c_1$ \\
\hline
Talking on the phone & Talking on the phone & \cmark & $c_2$ \\
\hline
Messaging & Messaging & \cmark & $c_3$ \\
\hline
Talking with passengers & Talking with passengers & \cmark & $c_4$ \\
\hline
Reaching behind & Reaching behind & \cmark & $c_5$ \\
\hline
Adjusting radio & Adjusting radio & \cmark & $c_6$ \\
\hline
Drinking & Drinking & \cmark & $c_7$ \\
\hline
\diagbox[width=12em, height=\line]{}{} & \textcolor{red}{Unseen} & \cmark & $c_8$ \\
\hline
\end{tabular}}
\caption{Labels utilized during training and testing. We transform the original binary classification problem into a multiclassification one by annotating the test set. For convenience, we combine some of them, so there are seven classes (one for driving attentively and six for NDRAs) in the training set and one additional named ``\textcolor{red}{unseen}'' in the test set to accommodate any new NDRAs.}
\label{tbl:new_labels}
\end{table}

\subsection{Model Structures}
\label{sec:3.2}

\begin{figure}[tp]
    \centering
    \includegraphics[width=\linewidth]{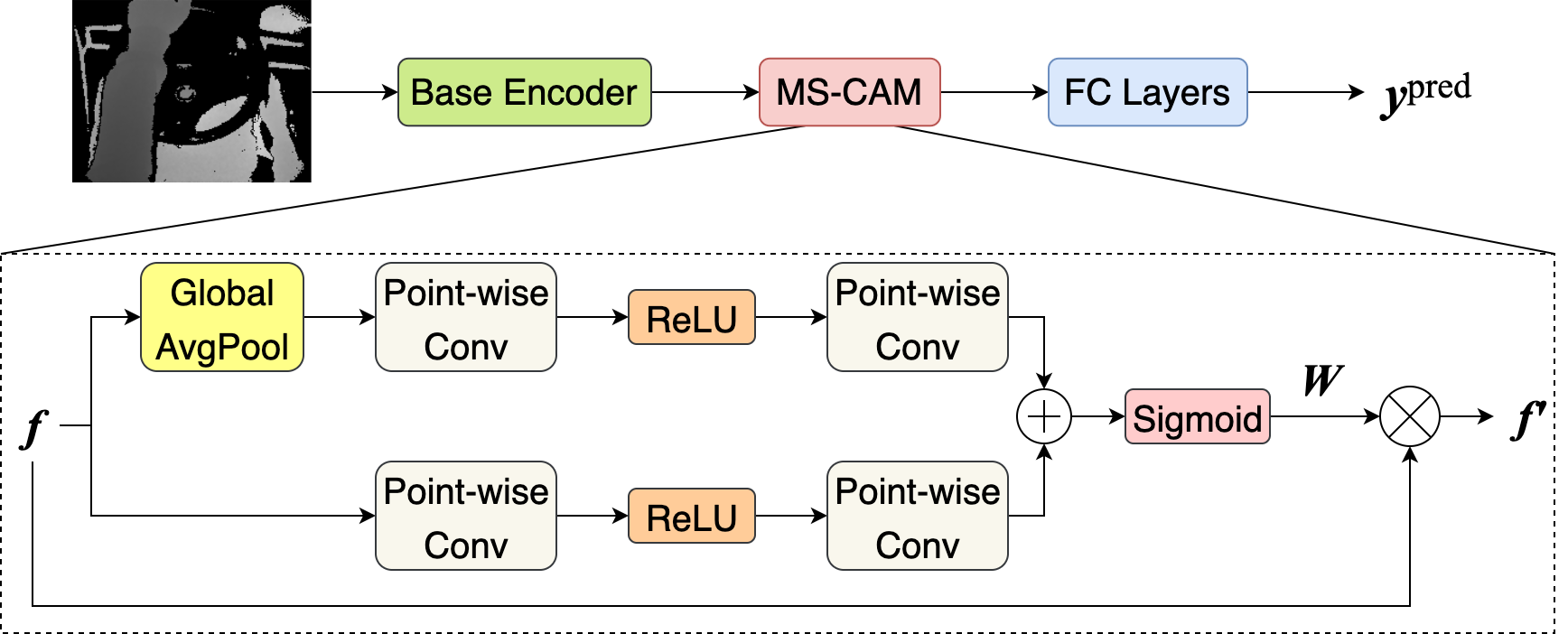}
    \caption{The architecture of our proposed unimodal DSM model. Pre-trained 2D CNNs are employed as the base encoders to extract features from the input image. Then the channel-wise attention is introduced into these features by leveraging the \textit{multi-scale channel attention module} (MS-CAM) \cite{dai2021attentional}. Finally, these features are passed to the \textit{fully connected} (FC) layers to make the prediction $\boldsymbol{y}^\text{pred}$.}
    \label{fig:single_mod}
\end{figure}

Assuming $\boldsymbol{X} \in \mathbb{R}^{C \times T \times H \times W}$ is a short video clip, where $T$ is the number of frames, $C$ indicates the number of channels, and $H$ and $W$ are the height and width, respectively. Our image-based models neglect the temporal dimension and consider only its last frame, i.e., $\boldsymbol{X}_{:, \, -1, \, :, \, :}$, which degenerates into an image. Following previous works \cite{baheti2018detection, kopuklu2021driver, ortega2020dmd}, we leverage pre-trained convolutional classifiers as the base encoder $\boldsymbol{\mathcal{E}}$ and \textit{fully connected neural networks} (FCNs) as the predition layers $\boldsymbol{\mathcal{P}}$.

\noindent \textbf{The unimodal case.} Given a single modality image $\boldsymbol{X}$, we exploit the base encoder to extract features
\begin{equation}
    \boldsymbol{f} = \boldsymbol{\mathcal{E}}(\boldsymbol{X}).
\end{equation}
Then, the \textit{multi-scale channel attention module} (MS-CAM) proposed in \cite{dai2021attentional} is leveraged to introduce global and local attention on $\boldsymbol{f}$. Based on the squeeze-and-excitation block \cite{hu2018squeeze}, it adds one more branch without the gobal pooling to preserve local information. Features from this two branches are fused via addition and then fed into the sigmoid function to generate weights $\boldsymbol{W}$. The features with attention $\boldsymbol{f}'$ are calculated by
\begin{equation}
    \boldsymbol{f}' = \boldsymbol{f} \otimes \boldsymbol{W}.
\end{equation}
where $\otimes$ stands for the element-wise multiplication. At last, $\boldsymbol{f}'$ is passed to $\boldsymbol{\mathcal{P}}$ to make the inference $\boldsymbol{y} = \boldsymbol{\mathcal{P}}(\boldsymbol{f}') \in \mathbb{R}^7$. The whole process is illustrated in Fig. \ref{fig:single_mod}.

\noindent\textbf{The multimodal scenario.} We utlise the cross-view and cross-modality case to illustrate our multimodal approach. Suppose $\boldsymbol{X}^{\text{T, IR}}$, $\boldsymbol{X}^{\text{T, D}}$, $\boldsymbol{X}^{\text{F, IR}}$ and $\boldsymbol{X}^{\text{F, D}}$ are four synchronized frames respectively from the top IR, the top depth, the front IR and the front depth cameras. Considering that these images have different modalities, we thus leverage four separate encoders $\boldsymbol{\mathcal{E}}^{\text{T, IR}}$, $\boldsymbol{\mathcal{E}}^{\text{T, D}}$, $\boldsymbol{\mathcal{E}}^{\text{F, IR}}$ and $\boldsymbol{\mathcal{E}}^{\text{F, D}}$ to correspondingly extract spatial features $\boldsymbol{f}^{\text{T, IR}}$, $\boldsymbol{f}^{\text{T, D}}$, $\boldsymbol{f}^{\text{F, IR}}$ and $\boldsymbol{f}^{\text{F, D}}$. To combine them together, there are two distinct avenues: feature-level fusion vs. decision-level fusion.

\begin{figure}[htp]
\begin{minipage}[t]{\linewidth}
    \centering
    \centerline{\includegraphics[width=\linewidth]{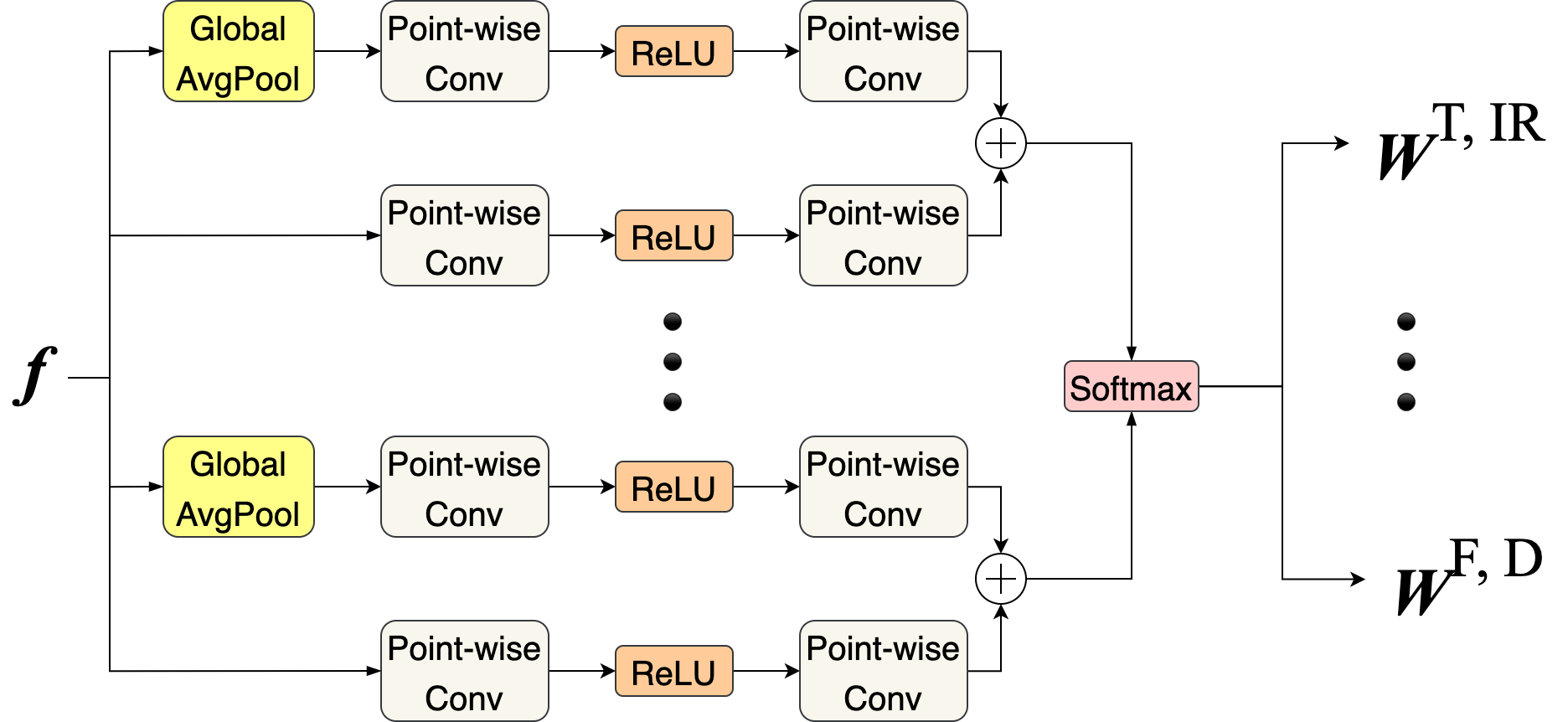}}
    \centerline{(a). MS-CAM.}\medskip
\end{minipage}
\begin{minipage}[t]{0.49\linewidth}
    \centering
    \centerline{\includegraphics[width=\linewidth]{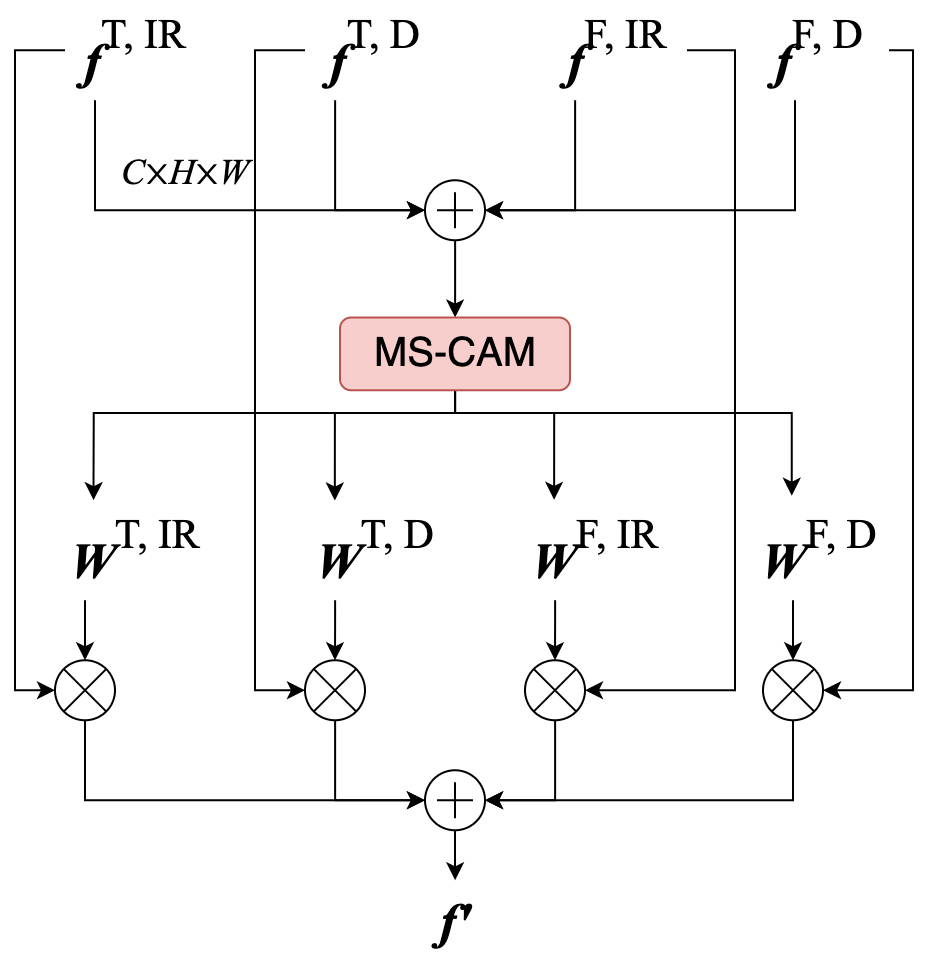}}
    \centerline{(b). AFF.} \medskip
\end{minipage}
\hfill
\begin{minipage}[t]{0.49\linewidth}
    \centering
    \centerline{\includegraphics[width=\linewidth]{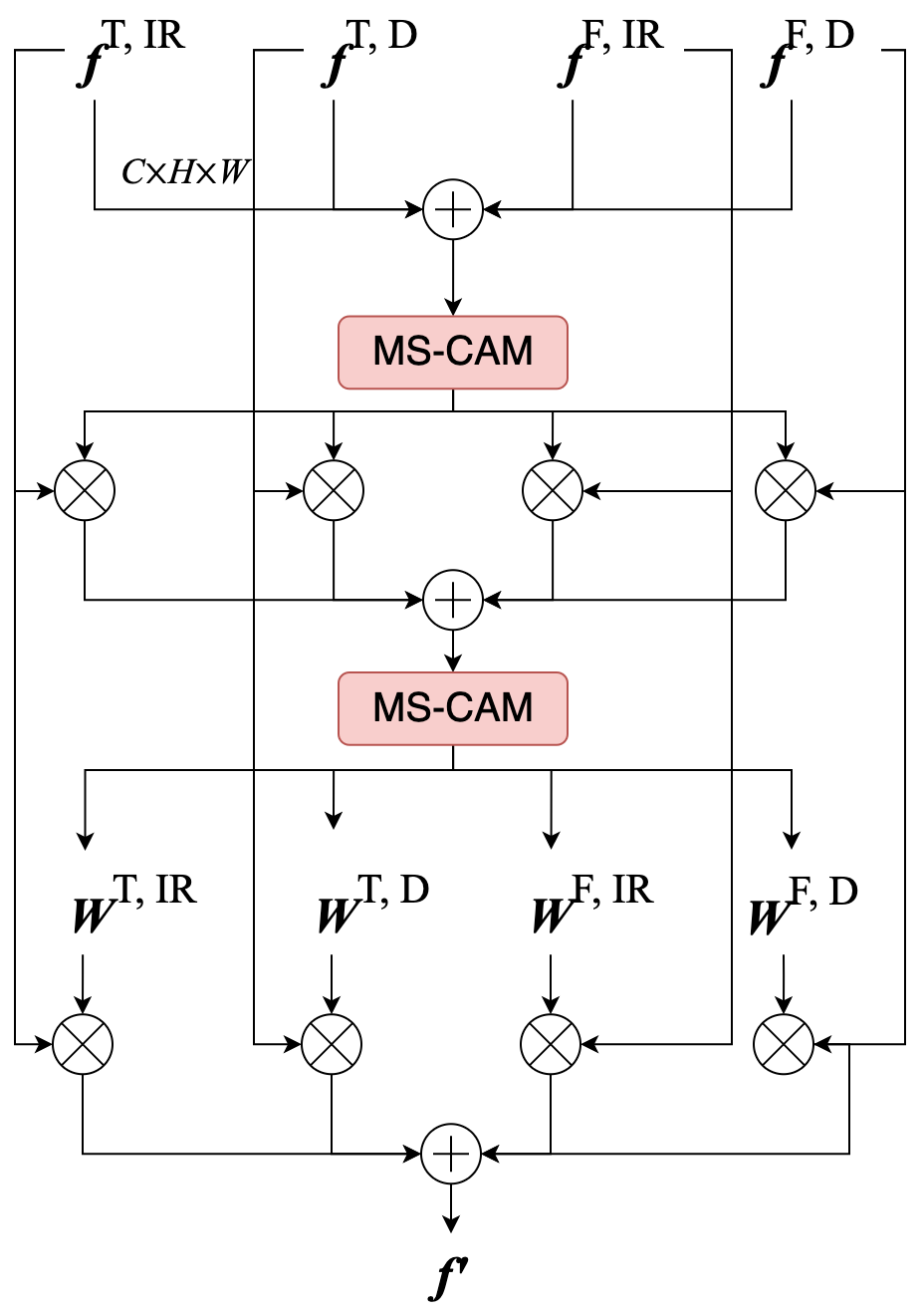}}
    \centerline{(c). iAFF.} \medskip
\end{minipage}
\caption{The architecture of our proposed multimodal fusion mechanism. Based on the original \textit{multi-scale channel attention module} (MS-CAM), \textit{attentional feature fusion} (AFF) and \textit{iterative attentional feature fusion} (iAFF) \cite{dai2021attentional}, we extend the number of attention head in MS-CAM to generate of four weight matrices, as illustrated by (a). Correspondingly, we replace the original MS-CAM in AFF and iAFF with our modified version to allow one-time fusion of four feature maps, as shown by (b) and (c).}
\label{fig:new_ms_cam}
\end{figure}

\begin{itemize}[itemsep=2pt,topsep=0pt,parsep=0pt]
    \item \textbf{Feature-level fusion}. The state-of-the-art feature fusion method AFF and iAFF proposed in \cite{dai2021attentional} can only combine two feature maps simultaneously, because their MS-CAM only contains one head, which can at most generate two weight matrices. For example, in Fig. \ref{fig:single_mod}, after acquiring one weight matrix $\boldsymbol{W}$, the other is determined by $\boldsymbol{1} - \boldsymbol{W}$. Hence, to enable the one-time fusion of features from four sources, we extend the number of heads in MS-CAM and use a softmax function to distribute weights, denoted by $\boldsymbol{W}^{\text{T, IR}}$ $\boldsymbol{W}^{\text{T, D}}$, $\boldsymbol{W}^{\text{F, IR}}$ and $\boldsymbol{W}^{\text{F, D}}$. The structure of our multi-head MS-CAM is illustrated in Figure \ref{fig:new_ms_cam}. Then the features fused by AFF are computed by
    \begin{equation}
        \boldsymbol{f}' = \sum_{\substack{i \in \{\text{T}, \, \text{F}\} \\ j \in \{\text{IR}, \, \text{D}\}}} \boldsymbol{f}^{i, \, j} \otimes \boldsymbol{W}^{i, \, j}.
    \end{equation}
    Finally, we feed $\boldsymbol{f}'$ to a shared FCN $\boldsymbol{\mathcal{P}}$ to make the prediction $\boldsymbol{y}$.
    \item \textbf{Decision-level fusion}. Similar to the single-modality case, we feed these features to four separate original MS-CAM \cite{dai2021attentional} to introduce global and local attention into each feature. Notice that the attention here is not related to views and modalities. Subsequently, these features are passed to four independent FCNs to calculate scores $\boldsymbol{y}^{\text{T, IR}}$ $\boldsymbol{y}^{\text{T, D}}$, $\boldsymbol{y}^{\text{F, IR}}$ and $\boldsymbol{y}^{\text{F, D}}$. In the last step, these scores are averaged to collectively make the final prediction:
    \begin{equation}
        \boldsymbol{y} = \frac{1}{4}\sum_{\substack{i \in \{\text{T}, \, \text{F}\} \\ j \in \{\text{IR}, \, \text{D}\}}} \boldsymbol{y}^{i, \, j}.
    \end{equation}
\end{itemize}

\begin{figure}[tp]
    \centering
    \includegraphics[width=\linewidth]{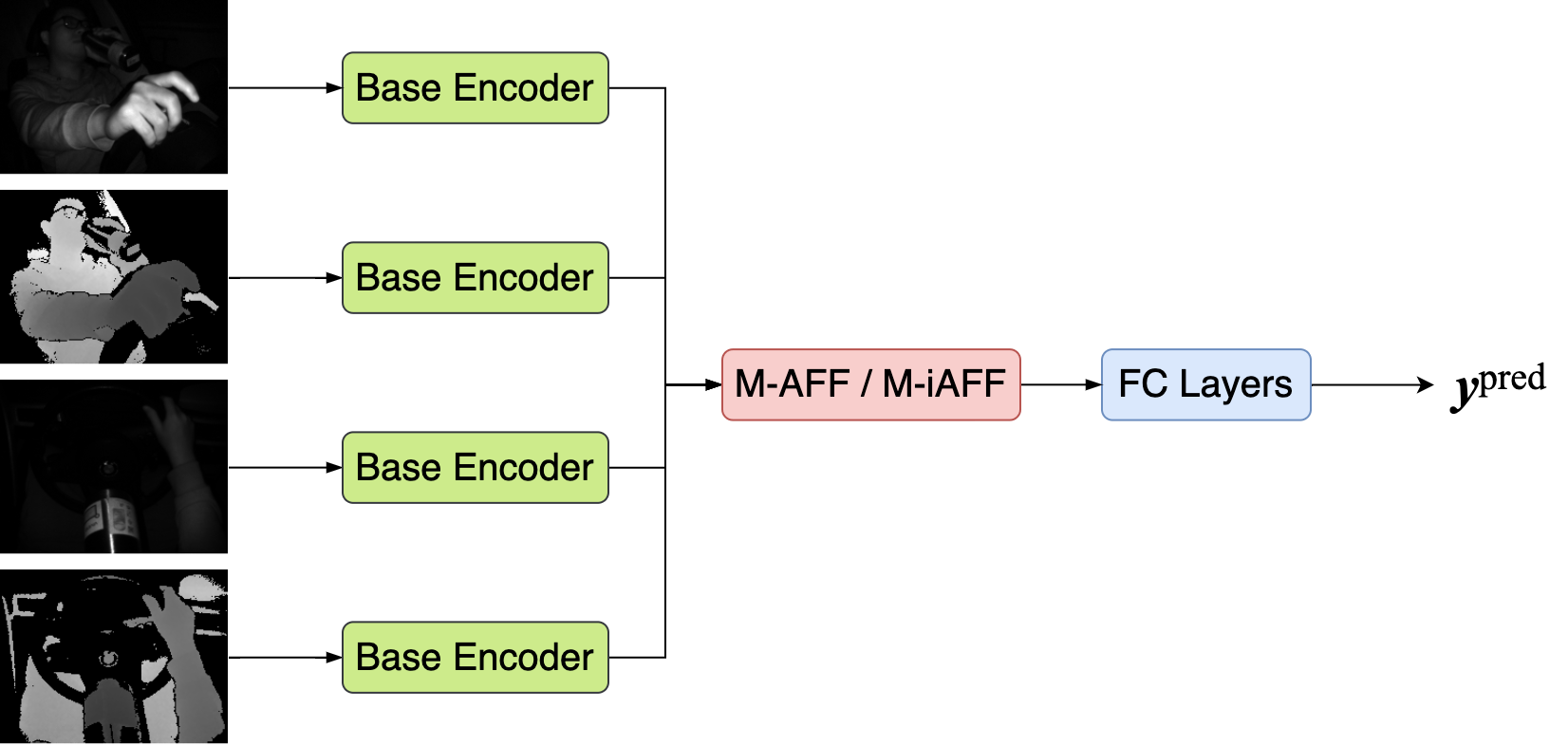}
    \caption{The architecture of our proposed multimodal DMS with feature-level fusion. Similar to the unimodal case, we employ pre-trained 2D CNNs as the base encoder. To fuse features with different modalities, we replace the original MS-CAM \cite{dai2021attentional} within AFF (or iAFF) block with our modified multi-head version (Fig. \ref{fig:new_ms_cam}). Then the fused feature is passed to the fully connected layers to predict the score.}
    \label{fig:multi_mod}
\end{figure}

\subsection{Testing Inferences}
\label{sec:3.3}

Since there are unseen classes in the test set, following \cite{kopuklu2021driver}, we utilize a threshold $\gamma$ to determine whether the predition should be ``unseen'' ($c_8$) or not. The prediction of the class label is determined by
\begin{equation}
y = 
\begin{cases}
    c_{\argmax \boldsymbol{y}}, \qquad &\text{if} \, \max \boldsymbol{y} \ge \gamma \\
    c_8, \qquad & \text{otherwise}
\end{cases}
\end{equation}
where $\max \boldsymbol{y}$ denotes the maximum element of $\boldsymbol{y}$.

\begin{table*}[tp]
\resizebox{\textwidth}{!}{
\begin{tabular}{l | c c c c | c c c c | c c c c}
\hline
\multirow{2}{*}{\textbf{Activity}} & \multicolumn{4}{c|}{\textbf{Histogram Similarity} $\uparrow$}                              & \multicolumn{4}{c|}{\textbf{Pixel-wise RMSE} $\downarrow$}                                   & \multicolumn{4}{c}{\textbf{Structural Similarity} $\uparrow$}                              \\ \cline{2-13} 
                                   & \textbf{Top IR} & \textbf{Top Depth} & \textbf{Front IR} & \textbf{Front Depth} & \textbf{Top IR} & \textbf{Top Depth} & \textbf{Front IR} & \textbf{Front Depth} & \textbf{Top IR} & \textbf{Top Depth} & \textbf{Front IR} & \textbf{Front Depth} \\ \hline
Normal driving                     & 0.99            & 1.00               & 0.97              & 1.00                 & 2.75            & 2.18               & 3.40              & 2.02                 & 0.96            & 0.84               & 0.95              & 0.88                 \\
talking on the phone               & 0.98            & 1.00               & 0.98              & 1.00                 & 2.01            & 1.02               & 2.98              & 1.82                 & 0.97            & 0.87               & 0.96              & 0.89                 \\
Messaging                          & 0.98            & 1.00               & 0.98              & 1.00                 & 2.21            & 2.01               & 2.82              & 1.78                 & 0.97            & 0.86               & 0.96              & 0.90                 \\
Talking with passengers            & 0.98            & 1.00               & 0.98              & 1.00                 & 2.06            & 1.97               & 3.14              & 1.87                 & 0.98            & 0.86               & 0.96              & 0.89                 \\
Reaching behind                    & 0.98            & 1.00               & 0.97              & 1.00                 & 2.05            & 2.02               & 3.46              & 1.90                 & 0.97            & 0.86               & 0.95              & 0.88                 \\
Adjusting radio                    & 0.99            & 1.00               & 0.96              & 1.00                 & 2.11            & 2.05               & 2.78              & 1.80                 & 0.97            & 0.86               & 0.96              & 0.89                 \\
Drinking                           & 0.99            & 1.00               & 0.98              & 1.00                 & 2.19            & 2.06               & 2.91              & 1.95                 & 0.98            & 0.86               & 0.92              & 0.88                 \\ \hline
\end{tabular}
}
\caption{Average similarities or (differences) under three distinct metrics. The arrows imply whether larger values indicate larger similarities ($\uparrow$) or not ($\downarrow$). The histogram and structural similarity should lie between 0 and 1, and being closer to 1 means high similarity. The pixel-wise RMSE is calculated based on the original pixel values in $[0, \, 255]$. The high resemblance of neighboring frames of all activities and modalities demonstrates that processing all frames within a video clip is unnecessary.}
\label{tbl:frame_sims}
\end{table*}

\section{Experiments}
\label{sec:4}

In this section, we first quantitatively prove the strong similarities between consecutive frames, which questions the essentiality of considering the temporal relations (Sec. \ref{sec:4.1}). Then, we demonstrate the effectiveness of our proposed approaches by comparing them with state-of-the-art methods \cite{kopuklu2021driver} (Sec. \ref{sec:4.2}). The results of multiclassification are shown and thoroughly analyzed in Sec. \ref{sec:4.3}. As for Sec. \ref{sec:4.4}, it illusrates the our models' real-time inference speeds compared with those in \cite{kopuklu2021driver}, and in Sec. \ref{sec:4.5}, the two fusion methods proposed in Sec. \ref{sec:3.2} are compared.

To make fair comparisons with the latest methods, we adopt the 2D versions of the base encoders used in \cite{kopuklu2021driver}, namely, 2D ResNet-18 \cite{he2016deep} and 2D MobileNet-V2 \cite{sandler2018mobilenetv2}. We also feed our models with video clips of 16 frames, with the same spatial data augmentation techniques applied in \cite{kopuklu2021driver}. We leverage the supervised contrastive loss \cite{khosla2020supervised} to fine-tune base encoders $\boldsymbol{\mathcal{E}}$ and the cross-entropy loss to train the prediction layers $\boldsymbol{\mathcal{P}}$. Each model is trained for 50 epochs with a batch size of 128. We employ the Adam algorithm \cite{kingma2014adam} with an initial learning rate of $1 \times 10^{-4}$ and a weight decay rate of $5 \times 10^{-3}$ as the optimizer. A cosine annealing schedule with warm restarts \cite{loshchilov2016sgdr} is exploited to tune the learning rate. Models are implemented in PyTorch and trained on a server with the Ubuntu 20.04 LTS OS and one NVIDIA RTX 3090.

\subsection{Frame Similarities}
\label{sec:4.1}

We adopt three common methods to measure how similar two neighboring frames are: the \textit{histogram similarity}, the \textit{mean squared of pixel-wise differences}, and the \textit{structural similarity} \cite{wang2004image}. The histogram similarity is based on the difference in the distribution of pixel values. The RMSE is a more rigid metric and can directly show the average pixel differences. The structural similarity is distinct from the former two metrics and is more global and similar to human perceptions. The results based on each activity and modality are shown in Table \ref{tbl:frame_sims}, from which we can see that consecutive frames are highly alike regardless of the choice of metrics. This observation compromises the necessity of considering the temporal dimension, and thus, exploiting 3D convolution on DAD \cite{kopuklu2021driver} is nonessential.

\subsection{Binary Classification}
\label{sec:4.2}

\begin{table*}[htbp]
\resizebox{\textwidth}{!}{
\begin{tabular}{l|ccccccccc}
\hline
\multirow{3}{*}{\textbf{Model}}       & \multicolumn{9}{c}{\textbf{AUC-ROC}  (\%)}                                                                                                                                                                               \\ \cline{2-10} 
                                      & \multicolumn{3}{c|}{\textbf{Top}}                                         & \multicolumn{3}{c|}{\textbf{Front}}                                       & \multicolumn{3}{c}{\textbf{Top + Front}}                        \\ \cline{2-10} 
                                      & \textbf{Depth} & \textbf{IR}   & \multicolumn{1}{c|}{\textbf{Depth + IR}} & \textbf{Depth} & \textbf{IR}   & \multicolumn{1}{c|}{\textbf{Depth + IR}} & \textbf{Depth}      & \textbf{IR}         & \textbf{Depth + IR} \\ \hline
MobileNet-V2 \cite{kopuklu2021driver} & 91.2           & 85.3          & \multicolumn{1}{c|}{91.5}                & 89.0           & 83.6          & \multicolumn{1}{c|}{89.8}                & {\ul \textbf{96.4}} & 91.5                & 96.1                \\
ResNet-18 \cite{kopuklu2021driver}    & 91.3           & 88.0          & \multicolumn{1}{c|}{91.7}                & 90.0           & 87.0          & \multicolumn{1}{c|}{\textbf{92.0}}       & 96.1                & 93.2                & {\ul \textbf{96.6}} \\ \hline
ours (MobileNet-V2 based)                   & \textbf{94.7}  & \textbf{94.9} & \multicolumn{1}{c|}{\textbf{94.8}}       & \textbf{92.0}  & 86.4          & \multicolumn{1}{c|}{90.0}                & 95.4                & {\ul \textbf{95.6}} & 93.6                \\
ours (ResNet-18 based)                      & 94.2           & 94.8          & \multicolumn{1}{c|}{94.2}                & 88.2           & \textbf{88.5} & \multicolumn{1}{c|}{90.4}                & 93.4                & 94.5                & {\ul 95.2}          \\ \hline
\end{tabular}
}
\caption{Comparison of models under AUC-ROC in binary classification. In each column (modality), models with the best scores in are indicated in bold typeface, and for each model, the scores of the best modalities are underlined.}
\label{tbl:auc_roc}
\end{table*}

\begin{table*}[tp]
\resizebox{\linewidth}{!}{
\begin{tabular}{l|ccccccccc}
\hline
\multirow{3}{*}{\textbf{Model}}    & \multicolumn{9}{c}{\textbf{Accuracy} (\%)}                                                                                                                                                                          \\ \cline{2-10} 
                                   & \multicolumn{3}{c|}{\textbf{Top}}                                               & \multicolumn{3}{c|}{\textbf{Front}}                                       & \multicolumn{3}{c}{\textbf{Top + Front}}             \\ \cline{2-10} 
                                   & \textbf{Depth} & \textbf{IR}         & \multicolumn{1}{c|}{\textbf{Depth + IR}} & \textbf{Depth} & \textbf{IR}   & \multicolumn{1}{c|}{\textbf{Depth + IR}} & \textbf{Depth} & \textbf{IR}   & \textbf{Depth + IR} \\ \hline
ResNet-18 \cite{kopuklu2021driver} & 89.1           & 83.6                & \multicolumn{1}{c|}{87.8}                & 87.2           & 83.7          & \multicolumn{1}{c|}{\textbf{88.7}}       & \textbf{91.6}  & 87.1          & {\ul \textbf{92.3}} \\ \hline
Ours (MobileNet-V2 based)                & \textbf{91.4}  & {\ul \textbf{92.4}} & \multicolumn{1}{c|}{91.7}                & \textbf{87.6}  & 81.5          & \multicolumn{1}{c|}{85.0}                & 90.4           & \textbf{90.0} & 89.2                \\
Ours (ResNet-18 based)                   & 91.2           & {\ul 91.9}          & \multicolumn{1}{c|}{{\ul \textbf{91.9}}} & 85.0           & \textbf{84.4} & \multicolumn{1}{c|}{86.7}                & 88.9           & 88.4          & 90.0                \\ \hline
\end{tabular}
}
\caption{Comparison of models in terms of binary classification accuracy. The highest score of each column is indicated in bold and that for each row is also underlined.}
\label{tbl:accuracy}
\end{table*}

The work \cite{kopuklu2021driver} is based on binary classification (distinguishing between $c_0$ and $\{c_1, \, \cdots, \, c_8\}$). We first compare our models with theirs under the \textit{area under the receiver-operating characteristic curve} ({AUC-ROC}) and the {accuracy}, which are utilized in \cite{kopuklu2021driver}. Since the test data is imbalanced (66.2\% ``normal driving'' and 33.8\% NDRAs), we also leverage \textit{the area under the precision-recall curve} ({AUC-PR}) as a more unbiased metric. We adopt feature-level fusion to handle multimodality. Results in terms of AUC-ROC, accuracy, and AUC-PR are shown in Table \ref{tbl:auc_roc}, Table \ref{tbl:accuracy} and Table \ref{tbl:auc_pr}, respectively. They confirm that our models can achieve state-of-the-art performance on the DAD \cite{kopuklu2021driver} dataset (AUC-ROC: 95.6 \% vs. 96.6\%; accuracy: 92.4\% vs. 92.3\%).

Table \ref{tbl:auc_roc} and Table \ref{tbl:accuracy} demonstrate that our image-based models can outperform their video-based counterparts in most cases, when using the top view and particularly  with the IR channel. In the case of the front view, 3D CNNs are slightly superior to ours only when IR and depth are fused. For the case of combining the top and front views, our proposed methods are again predominant with the IR modality.
The best performance of all four models, in terms of AUC-ROC, is attained when fusing the top view and the front view, while in terms of accuracy, our models have the best performances whe using the top view with IR data (92.4\% \& 91.9\%).

Table \ref{tbl:auc_pr} compares the AUC-PR values of our approaches. They cannot overall dominate one another in all modalities. MobileNet-V2, with the front and top views and the IR channel, achieves the highest score (97.5\%), while the best score (97.1\%) of ResNet-18 is reached when incorporating all modalities.

\begin{table}[tp]
\resizebox{\linewidth}{!}{
\begin{tabular}{l|cc}
\hline
\multirow{2}{*}{\textbf{Modality}} & \multicolumn{2}{c}{\textbf{AUC-PR}  (\%)} \\ \cline{2-3} 
                                   & MobileNet-V2 Based  & ResNet-18 Based  \\ \hline
Top Depth                          & 95.7                 & \textbf{95.8}     \\
Top IR                             & 95.9                 & \textbf{96.0}     \\
Top (Depth + IR)                   & \textbf{96.4}        & 95.7              \\
Front Depth                        & \textbf{94.3}        & 91.9              \\
Front IR                           & 91.6                 & \textbf{91.9}     \\
Front (Depth + IR)                 & \textbf{93.9}        & 93.0              \\
(Top + Front) Depth                & \textbf{97.1}        & 95.9              \\
(Top + Front) IR                   & {\ul \textbf{97.5}}        & 96.7              \\
(Top + Front) (Depth + IR)         & 96.2                 & {\ul \textbf{97.1}}     \\ \hline
\end{tabular}
}
\caption{Comparison of our image-based models under the AUC-PR metric in binary classification. The scores of the best modalities are underlined, and for each given modality, the score of the best model is in shown in bold.}
\label{tbl:auc_pr}
\end{table}

\subsection{Multiclass Classification}
\label{sec:4.3}

\begin{figure*}[tp]
    \centering
    \includegraphics[width=\textwidth]{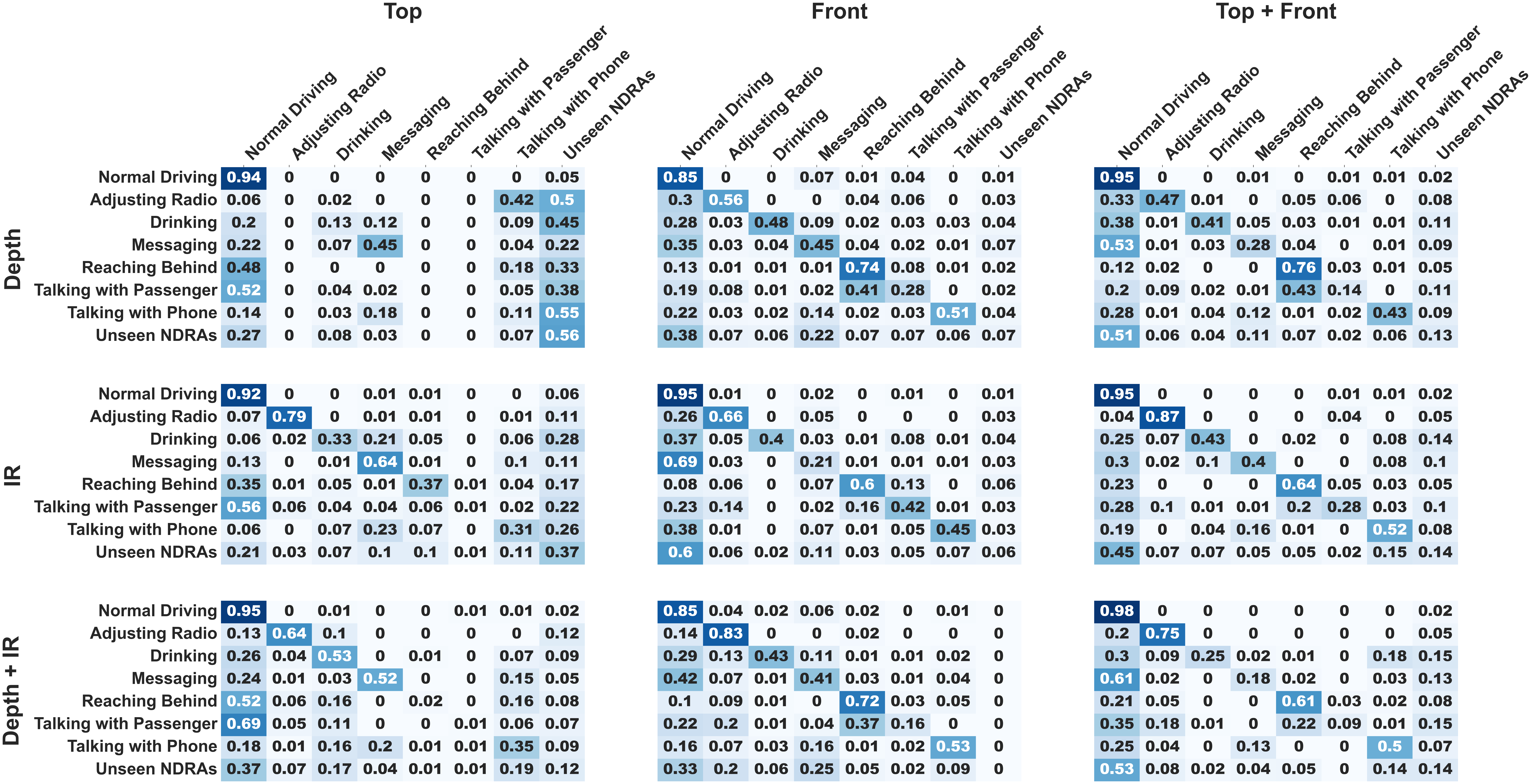}
    \caption{The confusion matrices of our ResNet-18-based model on DAD. There are $3 \times 3$ plots in this figure. Each row corresponds to a stream, and each column represents a view. The row of each confusion matrix is the ground-truth, while the column of each matrix is the prediction. Matrices are normalized to have  rows summing up to 1, hence diagnal entries represent the recall values.}
    \label{fig:confusion_mat}
\end{figure*}

Since the classes of DAD \cite{kopuklu2021driver} are unbalancedly distributed, we sample it during each epoch to ensure that each class has a similar number of instances. We leverage the method illustrated in Sec. \ref{sec:3.1} to predict the unseen class. Results in Table \ref{tbl:mult_acc} show that for our MobileNet-V2 \cite{sandler2018mobilenetv2} based approach, similar to the binary clase, it still achieve the highest multiclassification accuracy from the top view and the IR modality. However, the DMS based on ResNet-18 \cite{he2016deep} benefits more when combining the front and the top views. We visualize the confusion matrices of our ResNet-18-based model \cite{he2016deep} in Fig. \ref{fig:confusion_mat}. The following are important observations:

\begin{table}[tp]
\resizebox{\linewidth}{!}{
\begin{tabular}{l|cc}
\hline
\multirow{2}{*}{\textbf{Modality}} & \multicolumn{2}{c}{\textbf{Classification accuracy (\%)}} \\ \cline{2-3} 
                                   & MobileNet-V2 based        & ResNet-18 based            \\ \hline
Top Depth                          & \textbf{72.8}               & 72.7                        \\
Top IR                             & {\ul \textbf{75.0}}         & 74.5                        \\
Top (Depth + IR)                   & \textbf{73.7}               & 73.2                        \\
Front Depth                        & \textbf{72.4}               & 67.9                        \\
Front IR                           & 69.1                        & \textbf{73.3}               \\
Front (Depth + IR)                 & \textbf{71.1}               & 67.4                        \\
(Top + Front) Depth                & \textbf{72.9}               & \textbf{72.9}               \\
(Top + Front) IR                   & 72.0                        & {\ul \textbf{75.7}}         \\
(Top + Front) (Depth + IR)         & 73.0                        & \textbf{74.9}               \\ \hline
\end{tabular}
}
\caption{Comparison of multiclassification accuracies for different combination of modalities and views. The scores of the best modalities are underlined, and for each given modality, the score of the best model is in bold.}
\label{tbl:mult_acc}
\end{table}

\begin{itemize}[itemsep=2pt,topsep=0pt,parsep=0pt]
    \item \textit{Normal driving}. Our models can adequately recognize this activity in most cases. The highest recall value (98\%) is reached when fusing all four modalities.
    \item \textit{Adjusting radio}. The model based on Top \& Depth fails to recognize this action and confuses it with ``talking with phone'' and ``unseen'' NDRAs. By comparison, the model based on Top \& IR performs much better (79\%). The best score (87\%) is reached by the model based on (Top + Front) with IR. For the front view, fusing modalities can lead to better results, and similarly, for the IR modality, the combination of views provides higher scores.
    \item \textit{Drinking}. The model based on Top \& Depth  also fails to recognize the ``drinking'' activity with sufficient accuracy and misclassifies a significant proportion (45\%) as ``unseen''. By comparison, when also incorporating the IR modality, the results can be much better improved, with the highest recall score of 53\%. For other cases, combining views or modalities may not enhance the performance.
    \item \textit{Messaging}. This is an interesting case, as we find that except for Top \& IR, models based on other modalities are confused between ``messaging''  and ``normal driving''. This confusion represents a critical noise factor that could potentially cause severe consequences. Another finding is that the top view is more informative in recoginizing this action.
    \item \textit{Reaching behind}. Again, the model based on Top \& Depth  cannot correctly classify this action. This action can be better observed from the front view, which is intuitive. The largest recall value (76\%) is obtained when using (Top + Front) \& Depth.
    \item \textit{Talking with passenger}. Similar to ``reaching behind'' and as shown in Fig. \ref{fig:dad_samples}, this activity is only recognizable from the front view, but the front view model, unfortunately, achieves a recall score of only (42\%).  We belive the reason for this substandard performance is that the camera resolution is not high enough to capture the lip movements.
    \item \textit{Talking with phone}. Some models confuse this actitivity with ``normal driving'' and ``unseen'' NDRAs. Although the model based on Top \& IR  does not have the highest recall value, it appears to be capable to distinguish between this action and ``normal driving'' reasonably well.
    \item \textit{Unseen}. Models based on Top \& IR and Top \& Depth have more nontrival values (56\% and 37\%, respectively), while others seem uncapable to recognize this class well.
\end{itemize}
To summarize, although our models perform very well on the binary classification task, they cannot achieve similar performance on multiclassification. Among all activities, our models can recognize ``normal driving'' well but can be confused with other classes. We attribute the the low recall values of NDRAs to the limited representation power of ResNet-18 \cite{he2016deep} and the extreme imbalanced class distribution in the training set of DAD \cite{kopuklu2021driver}. On the other hand, we think overall the top view and the IR modality is more informative, while by contrast, the models based on Top \& Depth have suboptimal performances.

\subsection{Computational Efficiency}
\label{sec:4.4}

\begin{table}[tp]
\resizebox{\linewidth}{!}{
\begin{tabular}{l|cccc}
\hline
                                 & \multicolumn{4}{c}{\textbf{One Modality}}                                                                                                                               \\ \cline{2-5} 
\multirow{-2}{*}{\textbf{Model}} & \multicolumn{1}{c|}{\textbf{FLOPs}}    & \textbf{ARM64 (CPU)}                     & \textbf{ARM64 (GPU)}                     & \textbf{AMD64}                           \\ \hline
MobileNet-V2 \cite{kopuklu2021driver}                    & \multicolumn{1}{c|}{1.70 G}            & {\color[HTML]{009901} 0.2154 s}          & -                                        & {\color[HTML]{009901} 0.2157 s}          \\
ResNet-18 \cite{kopuklu2021driver}                       & \multicolumn{1}{c|}{24.05 G}           & {\color[HTML]{009901} 0.1881 s}          & -                                        & {\color[HTML]{009901} 0.0907 s}          \\ \hline
MobileNet-V2 (ours)              & \multicolumn{1}{c|}{\textbf{243.41 M}} & {\color[HTML]{009901} 0.0336 s}          & {\color[HTML]{009901} 0.0177 s}          & {\color[HTML]{009901} \textbf{0.0076 s}} \\
ResNet-18 (ours)                 & \multicolumn{1}{c|}{1.38 G}            & {\color[HTML]{009901} \textbf{0.0148 s}} & {\color[HTML]{009901} \textbf{0.0081 s}} & {\color[HTML]{009901} 0.0077 s}          \\ \hline
                                 & \multicolumn{4}{c}{\textbf{Two Modalities}}                                                                                                                             \\ \cline{2-5} 
\multirow{-2}{*}{\textbf{Model}} & \multicolumn{1}{c|}{\textbf{FLOPs}}    & \textbf{ARM64 (CPU)}                     & \textbf{ARM64 (GPU)}                     & \textbf{AMD64}                           \\ \hline
MobileNet-V2  \cite{kopuklu2021driver}                   & \multicolumn{1}{c|}{3.40 G}            & {\color[HTML]{CB0000} 0.4246 s}          & -                                        & {\color[HTML]{CB0000} 0.4393 s}          \\
ResNet-18  \cite{kopuklu2021driver}                      & \multicolumn{1}{c|}{48.09 G}           & {\color[HTML]{CB0000} 0.4328 s}          & -                                        & {\color[HTML]{009901} 0.1958 s}          \\ \hline
MobileNet-V2 (ours)              & \multicolumn{1}{c|}{\textbf{503.90 M}} & {\color[HTML]{009901} 0.0654 s}          & {\color[HTML]{009901} 0.0384s}           & {\color[HTML]{009901} 0.0163 s}          \\
ResNet-18 (ours)                 & \multicolumn{1}{c|}{2.77 G}            & {\color[HTML]{009901} \textbf{0.0294 s}} & {\color[HTML]{009901} \textbf{0.0195 s}} & {\color[HTML]{009901} \textbf{0.0157 s}} \\ \hline
                                 & \multicolumn{4}{c}{\textbf{Four Modalities}}                                                                                                                            \\ \cline{2-5} 
\multirow{-2}{*}{\textbf{Model}} & \multicolumn{1}{c|}{\textbf{FLOPs}}    & \textbf{ARM64 (CPU)}                     & \textbf{ARM64 (GPU)}                     & \textbf{AMD64}                           \\ \hline
MobileNet-V2  \cite{kopuklu2021driver}                   & \multicolumn{1}{c|}{6.81 G}            & {\color[HTML]{CB0000} 0.8456 s}          & -                                        & {\color[HTML]{CB0000} 0.9003 s}          \\
ResNet-18  \cite{kopuklu2021driver}                      & \multicolumn{1}{c|}{96.18 G}           & {\color[HTML]{CB0000} 0.7490 s}          & -                                        & {\color[HTML]{CB0000} 0.4051 s}          \\ \hline
MobileNet-V2 (ours)              & \multicolumn{1}{c|}{\textbf{1.01 G}}   & {\color[HTML]{009901} 0.1199 s}          & {\color[HTML]{009901} 0.0754 s}          & {\color[HTML]{009901} 0.0315 s}          \\
ResNet-18 (ours)                 & \multicolumn{1}{c|}{5.55 G}            & {\color[HTML]{009901} \textbf{0.0529 s}} & {\color[HTML]{009901} \textbf{0.0376 s}} & {\color[HTML]{009901} \textbf{0.0306 s}} \\ \hline
\end{tabular}
}
\caption{Comparison of models' efficiencies in terms of floating-point operations and inference times. {\color[HTML]{009901}Green} numbers indicate that models are rea-time while those in {\color[HTML]{CB0000}red} are not.}
\label{tbl:efficiency}
\end{table}

We compare our models with those leveraged in \cite{kopuklu2021driver} in terms of the number of \textbf{floating-point operations} (\textbf{FLOPs}) and the inference time. Each model is tested 1,000 times to reduce the effect of stochasticity. During each test-run, a randomly generated matrix of size $(16, \, 171, \, 224)$, which simulates a normlized video clip, is fed to each model. We select two platforms to represent different scenarios:
\begin{itemize}[itemsep=2pt,topsep=0pt,parsep=0pt]
    \item An M1 Pro chip (ARM64) with an 8-core CPU ($6 \times 3.2$ GHz + $2 \times 2.06$ GHz) and a 14-core GPU (1.30 GHz), and a 16 GB RAM to simulate embedded devices in cars.
    \item An AMD 5900x (AMD64) with a 12-core CPU (3.7 GHz) and a 64 GB RAM to simulate cloud servers.
\end{itemize}
Code is implemented in Python 3.9.13 and PyTorch 1.12.1. Note that the Infineon CamBoard pico flexx cameras used during data collection in DAD \cite{kopuklu2021driver} have a frame rate of 45 Hz. A model should then complete the inference before the next video clip is collected and fed to it. Therefore, real-time models should take no longer than $16 / 45 \approx 0.3556$ seconds to infer each video clip. Table \ref{tbl:efficiency} presents a detailed comparison of models. From it we can see that all our image-based models can make predictions within the expected time and are 10 times faster than those that use 3D CNNs, demonstrating exceptional efficiencies.

\subsection{Comparison of Fusion Strategies}
\label{sec:4.5}

To handle multiple modalities, we utilize mainly two fusion mechanisms: feature-level fusion and decision-level fusion, as elaborated in Sec. \ref{sec:3.2}. Table \ref{tbl:fusion} compares them with the base encoder ResNet-18 \cite{he2016deep}, from which we can see that feature-level fusion outperforms decision level fusion when fusing two views. As for other cases, performance gaps are not significant.

\begin{table}[htbp]
\resizebox{\linewidth}{!}{
\begin{tabular}{l|cc}
\hline
\multirow{2}{*}{\textbf{Modality}}  & \multicolumn{2}{c}{\textbf{AUC-PR} (\%)}              \\ \cline{2-3} 
                                    & \textbf{Feature Level} & \textbf{Decision Level} \\ \hline
Top (Depth + IR)          & 95.7                   & \textbf{96.2}           \\
Front (Depth + IR)         & 93.0                   & \textbf{93.4}           \\
(Top + Front) Depth        & \textbf{95.9}          & 95.6                    \\
(Top + Front) IR           & \textbf{96.7}          & 95.1                    \\
(Top + Front) (Depth + IR) & \textbf{97.1}          & 95.7                    \\ \hline
\end{tabular}}
\caption{The AUC-PR values of the two fusion methods with base encoder ResNet-18 \cite{he2016deep}. For each modality, the higher score is in bold.}
\label{tbl:fusion}
\end{table}

\section{Conclusion}

Based on our experiments, we discovered that compared with video-based models, our image-based methods can achieve similar performances and are exceedingly more efficient. Therefore, we can conclude that employing image-based models for the real-time inference of driver activities is more practical and less computationally complex. As part of this work, we also augmented the labelling of the DAD test set for future use by the scientific community. We found that data from the top view and the infrared stream are more informative and enhance model performance as confirmed by comparatively higher scores for binary classification and negligile confusion between phone-involved activities and normal driving. These results demosntrate a good practical value. Our models, however, do not perform the same on the multiclassification of NDRTs as on the binary classification problem. We ascribe this issue to the limited representation power of the backbones and to data imbalance. In the future, we plan to replace the backbone with the state-of-the-art classifiers and leverage more advanced techniques to imporve  multiclass classification performance.

{\small
\printbibliography

@inproceedings{kopuklu2021driver,
  title={Driver anomaly detection: A dataset and contrastive learning approach},
  author={K{\"o}p{\"u}kl{\"u}, Okan and Zheng, Jiapeng and Xu, Hang and Rigoll, Gerhard},
  booktitle={Proceedings of the IEEE/CVF Winter Conference on Applications of Computer Vision},
  pages={91--100},
  year={2021}
}

@inproceedings{dai2021attentional,
  title={Attentional feature fusion},
  author={Dai, Yimian and Gieseke, Fabian and Oehmcke, Stefan and Wu, Yiquan and Barnard, Kobus},
  booktitle={Proceedings of the IEEE/CVF Winter Conference on Applications of Computer Vision},
  pages={3560--3569},
  year={2021}
}

@article{ohn2013driver,
  title={Driver hand activity analysis in naturalistic driving studies: challenges, algorithms, and experimental studies},
  author={Ohn-Bar, Eshed and Martin, Sujitha and Trivedi, Mohan},
  journal={Journal of Electronic Imaging},
  volume={22},
  number={4},
  pages={041119},
  year={2013},
  publisher={SPIE}
}

@inproceedings{das2015performance,
  title={On performance evaluation of driver hand detection algorithms: Challenges, dataset, and metrics},
  author={Das, Nikhil and Ohn-Bar, Eshed and Trivedi, Mohan M},
  booktitle={2015 IEEE 18th international conference on intelligent transportation systems},
  pages={2953--2958},
  year={2015},
  organization={IEEE}
}

@inproceedings{kopuklu2020drivermhg,
  title={Drivermhg: A multi-modal dataset for dynamic recognition of driver micro hand gestures and a real-time recognition framework},
  author={K{\"o}p{\"u}kl{\"u}, Okan and Ledwon, Thomas and Rong, Yao and Kose, Neslihan and Rigoll, Gerhard},
  booktitle={2020 15th IEEE International Conference on Automatic Face and Gesture Recognition (FG 2020)},
  pages={77--84},
  year={2020},
  organization={IEEE}
}

@article{diaz2016reduced,
  title={A reduced feature set for driver head pose estimation},
  author={Diaz-Chito, Katerine and Hern{\'a}ndez-Sabat{\'e}, Aura and L{\'o}pez, Antonio M},
  journal={Applied Soft Computing},
  volume={45},
  pages={98--107},
  year={2016},
  publisher={Elsevier}
}

@inproceedings{roth2019dd,
  title={Dd-pose-a large-scale driver head pose benchmark},
  author={Roth, Markus and Gavrila, Dariu M},
  booktitle={2019 IEEE Intelligent Vehicles Symposium (IV)},
  pages={927--934},
  year={2019},
  organization={IEEE}
}

@inproceedings{schwarz2017driveahead,
  title={Driveahead-a large-scale driver head pose dataset},
  author={Schwarz, Anke and Haurilet, Monica and Martinez, Manuel and Stiefelhagen, Rainer},
  booktitle={Proceedings of the IEEE Conference on Computer Vision and Pattern Recognition Workshops},
  pages={1--10},
  year={2017}
}

@inproceedings{massoz2016ulg,
  title={The ULg multimodality drowsiness database (called DROZY) and examples of use},
  author={Massoz, Quentin and Langohr, Thomas and Fran{\c{c}}ois, Cl{\'e}mentine and Verly, Jacques G},
  booktitle={2016 IEEE Winter Conference on Applications of Computer Vision (WACV)},
  pages={1--7},
  year={2016},
  organization={IEEE}
}

@inproceedings{weng2016driver,
  title={Driver drowsiness detection via a hierarchical temporal deep belief network},
  author={Weng, Ching-Hua and Lai, Ying-Hsiu and Lai, Shang-Hong},
  booktitle={Asian Conference on Computer Vision},
  pages={117--133},
  year={2016},
  organization={Springer}
}

@inproceedings{abouelnaga2018real,
  title={Real-time Distracted Driver Posture Classification},
  author={Abouelnaga, Yehya and Eraqi, Hesham M and Moustafa, Mohamed N},
  booktitle={Neural Information Processing Systems (NIPS 2018), Workshop on Machine Learning for Intelligent Transportation Systems},
  year={2018}
}

@article{eraqi2019driver,
  title={Driver distraction identification with an ensemble of convolutional neural networks},
  author={Eraqi, Hesham M and Abouelnaga, Yehya and Saad, Mohamed H and Moustafa, Mohamed N},
  journal={Journal of Advanced Transportation},
  volume={2019},
  year={2019},
  publisher={Hindawi}
}

@inproceedings{martin2019drive,
  title={Drive\&act: A multi-modal dataset for fine-grained driver behavior recognition in autonomous vehicles},
  author={Martin, Manuel and Roitberg, Alina and Haurilet, Monica and Horne, Matthias and Rei{\ss}, Simon and Voit, Michael and Stiefelhagen, Rainer},
  booktitle={Proceedings of the IEEE/CVF International Conference on Computer Vision},
  pages={2801--2810},
  year={2019}
}

@inproceedings{ortega2020dmd,
  title={Dmd: A large-scale multi-modal driver monitoring dataset for attention and alertness analysis},
  author={Ortega, Juan Diego and Kose, Neslihan and Ca{\~n}as, Paola and Chao, Min-An and Unnervik, Alexander and Nieto, Marcos and Otaegui, Oihana and Salgado, Luis},
  booktitle={European Conference on Computer Vision},
  pages={387--405},
  year={2020},
  organization={Springer}
}

@inproceedings{baheti2018detection,
  title={Detection of distracted driver using convolutional neural network},
  author={Baheti, Bhakti and Gajre, Suhas and Talbar, Sanjay},
  booktitle={Proceedings of the IEEE conference on computer vision and pattern recognition workshops},
  pages={1032--1038},
  year={2018}
}

@inproceedings{simonyan2015very,
  author={Karen Simonyan and Andrew Zisserman},
  title={Very Deep Convolutional Networks for Large-Scale Image Recognition},
  booktitle={International Conference on Learning Representations},
  year={2015},
}

@inproceedings{carreira2017quo,
  title={Quo vadis, action recognition? a new model and the kinetics dataset},
  author={Carreira, Joao and Zisserman, Andrew},
  booktitle={proceedings of the IEEE Conference on Computer Vision and Pattern Recognition},
  pages={6299--6308},
  year={2017}
}

@inproceedings{kopuklu2019resource,
  title={Resource efficient 3d convolutional neural networks},
  author={K{\"o}p{\"u}kl{\"u}, Okan and Kose, Neslihan and Gunduz, Ahmet and Rigoll, Gerhard},
  booktitle={2019 IEEE/CVF International Conference on Computer Vision Workshop (ICCVW)},
  pages={1910--1919},
  year={2019},
  organization={IEEE}
}

@inproceedings{gutmann2010noise,
  title={Noise-contrastive estimation: A new estimation principle for unnormalized statistical models},
  author={Gutmann, Michael and Hyv{\"a}rinen, Aapo},
  booktitle={Proceedings of the thirteenth international conference on artificial intelligence and statistics},
  pages={297--304},
  year={2010},
  organization={JMLR Workshop and Conference Proceedings}
}

@inproceedings{tran2018closer,
  title={A closer look at spatiotemporal convolutions for action recognition},
  author={Tran, Du and Wang, Heng and Torresani, Lorenzo and Ray, Jamie and LeCun, Yann and Paluri, Manohar},
  booktitle={Proceedings of the IEEE conference on Computer Vision and Pattern Recognition},
  pages={6450--6459},
  year={2018}
}

@inproceedings{he2016deep,
  title={Deep residual learning for image recognition},
  author={He, Kaiming and Zhang, Xiangyu and Ren, Shaoqing and Sun, Jian},
  booktitle={Proceedings of the IEEE conference on computer vision and pattern recognition},
  pages={770--778},
  year={2016}
}

@inproceedings{sandler2018mobilenetv2,
  title={Mobilenetv2: Inverted residuals and linear bottlenecks},
  author={Sandler, Mark and Howard, Andrew and Zhu, Menglong and Zhmoginov, Andrey and Chen, Liang-Chieh},
  booktitle={Proceedings of the IEEE conference on computer vision and pattern recognition},
  pages={4510--4520},
  year={2018}
}

@article{wang2004image,
  title={Image quality assessment: from error visibility to structural similarity},
  author={Wang, Zhou and Bovik, Alan C and Sheikh, Hamid R and Simoncelli, Eero P},
  journal={IEEE transactions on image processing},
  volume={13},
  number={4},
  pages={600--612},
  year={2004},
  publisher={IEEE}
}

@inproceedings{hu2018squeeze,
  title={Squeeze-and-excitation networks},
  author={Hu, Jie and Shen, Li and Sun, Gang},
  booktitle={Proceedings of the IEEE conference on computer vision and pattern recognition},
  pages={7132--7141},
  year={2018}
}

@inproceedings{szegedy2015going,
  title={Going deeper with convolutions},
  author={Szegedy, Christian and Liu, Wei and Jia, Yangqing and Sermanet, Pierre and Reed, Scott and Anguelov, Dragomir and Erhan, Dumitru and Vanhoucke, Vincent and Rabinovich, Andrew},
  booktitle={Proceedings of the IEEE conference on computer vision and pattern recognition},
  pages={1--9},
  year={2015}
}

@inproceedings{szegedy2016rethinking,
  title={Rethinking the inception architecture for computer vision},
  author={Szegedy, Christian and Vanhoucke, Vincent and Ioffe, Sergey and Shlens, Jon and Wojna, Zbigniew},
  booktitle={Proceedings of the IEEE conference on computer vision and pattern recognition},
  pages={2818--2826},
  year={2016}
}

@inproceedings{szegedy2017inception,
  title={Inception-v4, inception-resnet and the impact of residual connections on learning},
  author={Szegedy, Christian and Ioffe, Sergey and Vanhoucke, Vincent and Alemi, Alexander A},
  booktitle={Thirty-first AAAI conference on artificial intelligence},
  year={2017}
}

@inproceedings{lin2017feature,
  title={Feature pyramid networks for object detection},
  author={Lin, Tsung-Yi and Doll{\'a}r, Piotr and Girshick, Ross and He, Kaiming and Hariharan, Bharath and Belongie, Serge},
  booktitle={Proceedings of the IEEE conference on computer vision and pattern recognition},
  pages={2117--2125},
  year={2017}
}

@inproceedings{ronneberger2015u,
  title={U-net: Convolutional networks for biomedical image segmentation},
  author={Ronneberger, Olaf and Fischer, Philipp and Brox, Thomas},
  booktitle={International Conference on Medical image computing and computer-assisted intervention},
  pages={234--241},
  year={2015},
  organization={Springer}
}

@inproceedings{li2019selective,
  title={Selective kernel networks},
  author={Li, Xiang and Wang, Wenhai and Hu, Xiaolin and Yang, Jian},
  booktitle={Proceedings of the IEEE/CVF conference on computer vision and pattern recognition},
  pages={510--519},
  year={2019}
}

@article{kingma2014adam,
  title={Adam: A method for stochastic optimization},
  author={Kingma, Diederik P and Ba, Jimmy},
  journal={arXiv preprint arXiv:1412.6980},
  year={2014}
}

@article{khosla2020supervised,
  title={Supervised contrastive learning},
  author={Khosla, Prannay and Teterwak, Piotr and Wang, Chen and Sarna, Aaron and Tian, Yonglong and Isola, Phillip and Maschinot, Aaron and Liu, Ce and Krishnan, Dilip},
  journal={Advances in Neural Information Processing Systems},
  volume={33},
  pages={18661--18673},
  year={2020}
}

@inproceedings{ma2018shufflenet,
  title={Shufflenet v2: Practical guidelines for efficient cnn architecture design},
  author={Ma, Ningning and Zhang, Xiangyu and Zheng, Hai-Tao and Sun, Jian},
  booktitle={Proceedings of the European conference on computer vision (ECCV)},
  pages={116--131},
  year={2018}
}

@article{loshchilov2016sgdr,
  title={Sgdr: Stochastic gradient descent with warm restarts},
  author={Loshchilov, Ilya and Hutter, Frank},
  journal={arXiv preprint arXiv:1608.03983},
  year={2016}
}
}

\end{document}